\pdfoutput=1
\documentclass[letterpaper, 10 pt, conference]{ieeeconf}  

\IEEEoverridecommandlockouts                              

\usepackage{graphicx} 
\usepackage{amsmath} 
\usepackage{cite}
\usepackage{array}
\newcolumntype{C}[1]{>{\centering\arraybackslash}m{#1}}
\usepackage[caption=false, font=footnotesize]{subfig}
\usepackage{siunitx}

\usepackage[dvipsnames, table]{xcolor}

\usepackage[bookmarks=true]{hyperref}

\title{Robotic Assembly across Multiple Contact Stiffnesses with\\ Robust
  Force Controllers }

\author{Lee Ying Jun Wilson$^{1}$ and Quang-Cuong Pham$^{1}$
  \thanks{$^{1}$ School of Mechanical and Aerospace
    Engineering, Nanyang Technological University, Singapore. Email:
    \texttt{cuong@ntu.edu.sg}.
        }%
}

\begin{document}

\maketitle
\thispagestyle{empty}
\pagestyle{empty}

\begin{abstract}


  Active Force Control (AFC) is an important scheme for tackling
  high-precision robotic assembly. Classical force controllers are
  highly surface-dependent: the controller must be carefully tuned for
  each type of surface in contact, in order to avoid instabilities and
  to achieve a reasonable performance level. Here, we build upon the
  recently-developed Convex Controller Synthesis (CCS) to enable
  high-precision assembly across a wide range of surface stiffnesses
  without any surface-dependent tuning. Specifically, we demonstrate
  peg-in-hole assembly with 100 micron clearance, initial position
  uncertainties up to 2 cm, and for four types of peg and hole
  materials -- rubber, plastic, wood, aluminum -- whose stiffnesses
  range from 10 to 100 N/mm, using a single controller.

\end{abstract}

\section{Introduction}
\label{section: introduction}

Active Force Control (AFC) is an important scheme for tackling
high-precision robotic assembly, such as peg or shaft insertion into
holes, memory module insertion onto motherboards, or gear assembly,
see e.g.~\cite{xu2019compare}. In AFC, contact force measurements
obtained from a wrist-mounted force sensor are used in a feedback loop
to guide the robot motion. A typical task in AFC is to maintain a
desired level of normal contact force between the robot and a
surface. Classical AFC suffers from the following fundamental problem:
the controller must be carefully tuned for each type of surface in
contact. A controller tuned for a soft surface might become unstable
when making contact with a hard surface, while a controller tuned for
a hard surface might be overly sluggish when the task involves soft
surfaces~\cite{pham2019convex}. Furthermore, the tuning of classical
AFC controllers is time-consuming and requires considerable
expertise. Because of those issues, high-precision automated assembly
under large uncertainties (in the surface stiffnesses, in the initial
positions of the parts) remains an outstanding industrial challenge.

In this paper, we build upon the recently-developed Convex Controller
Synthesis (CCS)~\cite{pham2019convex} to enable high-precision assembly
across a wide range of surface stiffnesses \emph{without any
  surface-dependent tuning}. Specifically, we demonstrate peg-in-hole
assembly with \SI{100}{\micro\metre} clearance, initial position
uncertainties up to \SI{2}{cm}, and for four types of peg and hole
materials -- rubber, plastic, wood, aluminum -- whose stiffnesses
range from 10 to \SI{100}{N/mm}, using a single controller.

Note that we use a position-controlled industrial robot in our
experiments. It is significantly more difficult to control contact
forces with position-controlled robots than with torque-controlled
robots. However, this also makes our results more widely applicable,
as the overwhelming majority of robots in the industry are
position-controlled, owing to their high precision and
cost-effectiveness~\cite{suarez2018can}.

This paper is organized as follows. Section~\ref{sec:background}
presents the background on AFC-based peg-in-hole
assembly. Section~\ref{sec:strategy} describes our high-level assembly
strategy. While the core of this high-level strategy is adapted from
the literature, we highlight a number of innovations that improve its
robustness, especially against variations in surface
stiffness. Section~\ref{sec:experiments} presents the peg-in-hole
assembly experiments using materials with a wide range of
stiffnesses. We show in particular that the proposed high-level
assembly strategy combined with a low-level CCS force controller
achieves outstanding performance and robustness on this challenging
task. Finally, Section~\ref{sec:conclusions} concludes and sketches
some directions for future research.

\section{Background}
\label{sec:background}

A typical peg-in-hole assembly pipeline comprises two elements: a
high-level assembly strategy, and a low-level force controller.

\subsection{High-level assembly strategy}

The classical high-level assembly strategy can be divided into two
phases: search phase and insertion phase. The purpose of the search
phase is to locate the hole plane (Z coordinate) and the hole position
(X-Y coordinates) within the plane. Typically, a hybrid position-force
control strategy is used to traverse the search plane (e.g. following
a spiral pattern) while maintaining a constant normal contact force
\cite{raibert1981hybrid}. The hole can then be located by tracking the
position of the end-effector.

Next, the insertion phase relies on active force control in all three
axes to ensure safe insertion without jamming or wedging
\cite{inoue1974force, newman2001interpretation,
  chhatpar2001search}. Whereas the rigid parts assembly has been
extensively researched and optimized, assembly strategies for flexible
parts appear less ventured \cite{bodenhagen2012learning}. For flexible
parts, contact deformation and stick-slip friction become apparent and
can jeopardize the successful localization of the hole
\cite{bodenhagen2012learning,bona2005friction}. Furthermore, unlike
rigid parts, the force dynamics of a flexible parts is non-linear and
more complex due to flexing
\cite{jasim2017contact,kim1999visual,yanchun2010assembly}. Hence, we
also develop a generic high-level force-controlled peg-in-hole
assembly strategy that is feasible for both rigid and flexible parts.

In addition to the above classical approaches, Reinforcement Learning
has been recently proposed to generate more flexible high-level
assembly strategies or components
thereof~\cite{inoue2017deep,luo2019reinforcement}.

\subsection{Low-level force controller}
\label{sec:controller}

Whether classical or learning-based, most high-level assembly
strategies must rely on a low-level force or impedance/admittance
controller to realize the desired behavior (e.g. tracking a constant
normal force, maintaining a desired impedance, etc.) 

\begin{figure}
\begin{center}
\includegraphics[width = 3.4in]{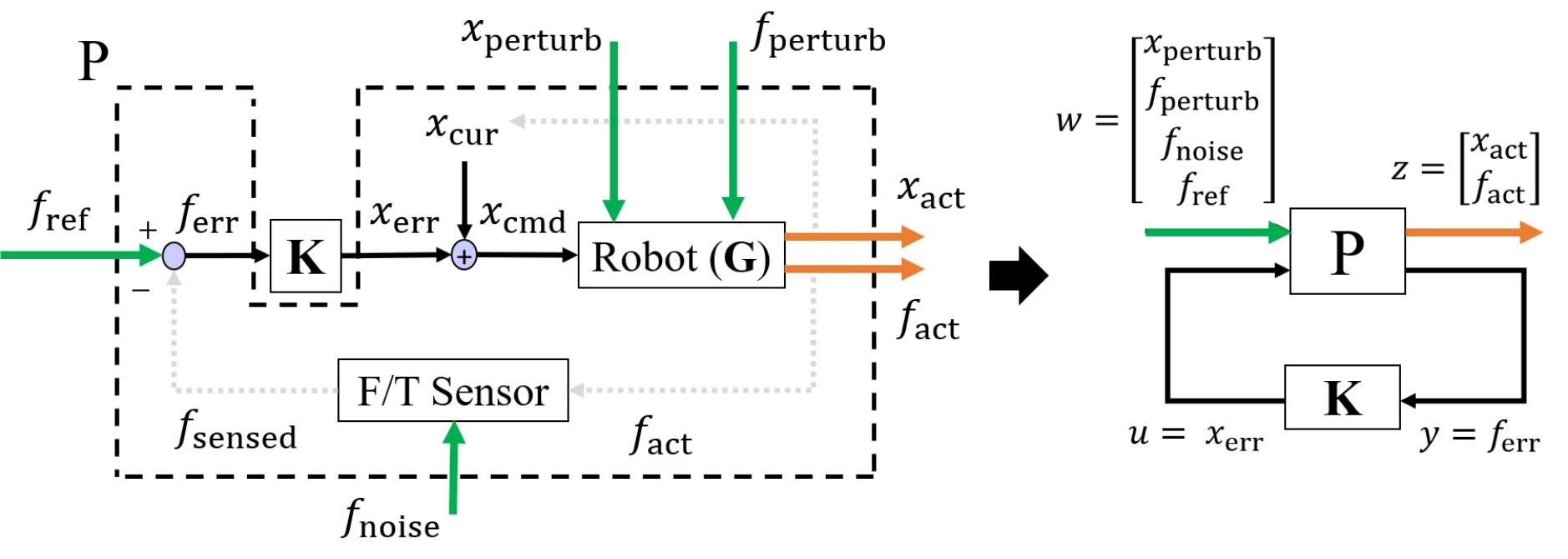}
\caption{General formulation of force control for a
  position-controlled industrial robot. Block $\mathbf{K}$ can
  represent a Proportional-Integral-Derivative (PID) controller, or
  one obtained by Convex Controller Synthesis (CCS).}
\label{fig:control}
\end{center}
\end{figure}

Consider the task of tracking a constant normal force by a
position-controlled industrial robot. As depicted in
Fig.~\ref{fig:control}, the force controller works to eliminate force
tracking errors, $f_{\mathrm{err}}$, by ``converting'' them into
position errors, $x_{\mathrm{err}}$, which are in turn tracked by the
robot's internal position controller. A well-developed force
controller should be able to quickly track a reference force without
going into instabilities. In this section, we discuss two low-level
force controllers, namely the PID and CCS force controllers.

\subsubsection{PID force controller}

The classical PID force controller is widely utilized in active force
control, as it is relatively simple to change the overall system
behavior in force tracking through each controller
gain~\cite{volpe1993theoretical,franklin1994feedback}. Depending on
the contact stiffness of the overall system, a variant of PID could
produce a feasible force controller $K$. It has been found that a pure
integral controller provides the best performance for force control,
with zero steady state error and a lower likelihood for instability
\cite{volpe1993theoretical}. This motivates the use of integral force
controller in our subsequent analysis. Yet, the performance of an
integral force controller is constrained by contact
stiffnesses. Tuning higher integral gains will attain better
performance in softer contacts but increases the risk of instability
under stiffer material contacts. On the other hand, having low
integral gains to accommodate stiffer contacts reduces the speed of
assembly and could lead to excessive force overshoot. For assemblies
requiring contact with multiple materials, overly conservative
integral gains will also lead to non-optimal performance on softer
materials.

\subsubsection{CCS force controller}

The CCS force controller is based on a recently-proposed controller
framework that searches for all stabilizing controllers to provide a
desired force tracking nominal performance with a robust stability to
a wide range of contact stiffnesses \cite{pham2019convex}.

Briefly, the force control task is formulated as a
multi-input-multi-output general control configuration using robust
control theory as illustrated in Fig.~\ref{fig:control}. A matrix
representation of the model could be as follows:
\[
\begin{bmatrix}
z \\ y
\end{bmatrix} = 
\begin{bmatrix}
P_{11} & P_{12} \\ P_{21} & P_{22}
\end{bmatrix} 
\begin{bmatrix}
w \\ u
\end{bmatrix},
\]
with $P_{11} - P_{22}$ derived from the system dynamics
\cite{skogestad2007multivariable}. Then, the linear fractional
transformation resulting in the closed-loop transfer matrix $H$ is:
\[
z = Hw, 
\]
\[
H = (P_{11} + P_{12}K)(I - P_{22}K)^{-1}P_{21}.
\]

Note that each element in $H$ corresponds to a transfer function from
the exogenous inputs $w$ to the exogenous outputs $z$. If we identify
our performance requirements, such as noise attenuation, we can
proceed to constrain and optimize each element to achieve our
performance requirements. The aim of the CCS framework is then to
search for the achievable closed-loop transfer matrix $H$ with the
lowest cost and thereafter, derive the appropriate stable controller
$K$.

Notably, this force controller is a fixed controller that is optimally
developed under constraints to achieve the desired nominal performance
with robust stability across a wide range of contact stiffnesses. This
implies that a CCS force controller, unlike classical PID controllers,
can be tuned to perform nominally well for a soft material and still
complete the assembly on stiffer materials at a similar assembly speed
without going into instability.

\section{Improvements to the high-level assembly strategy}
\label{sec:strategy}

\begin{figure}
\begin{center}
\includegraphics[width = 3.3in]{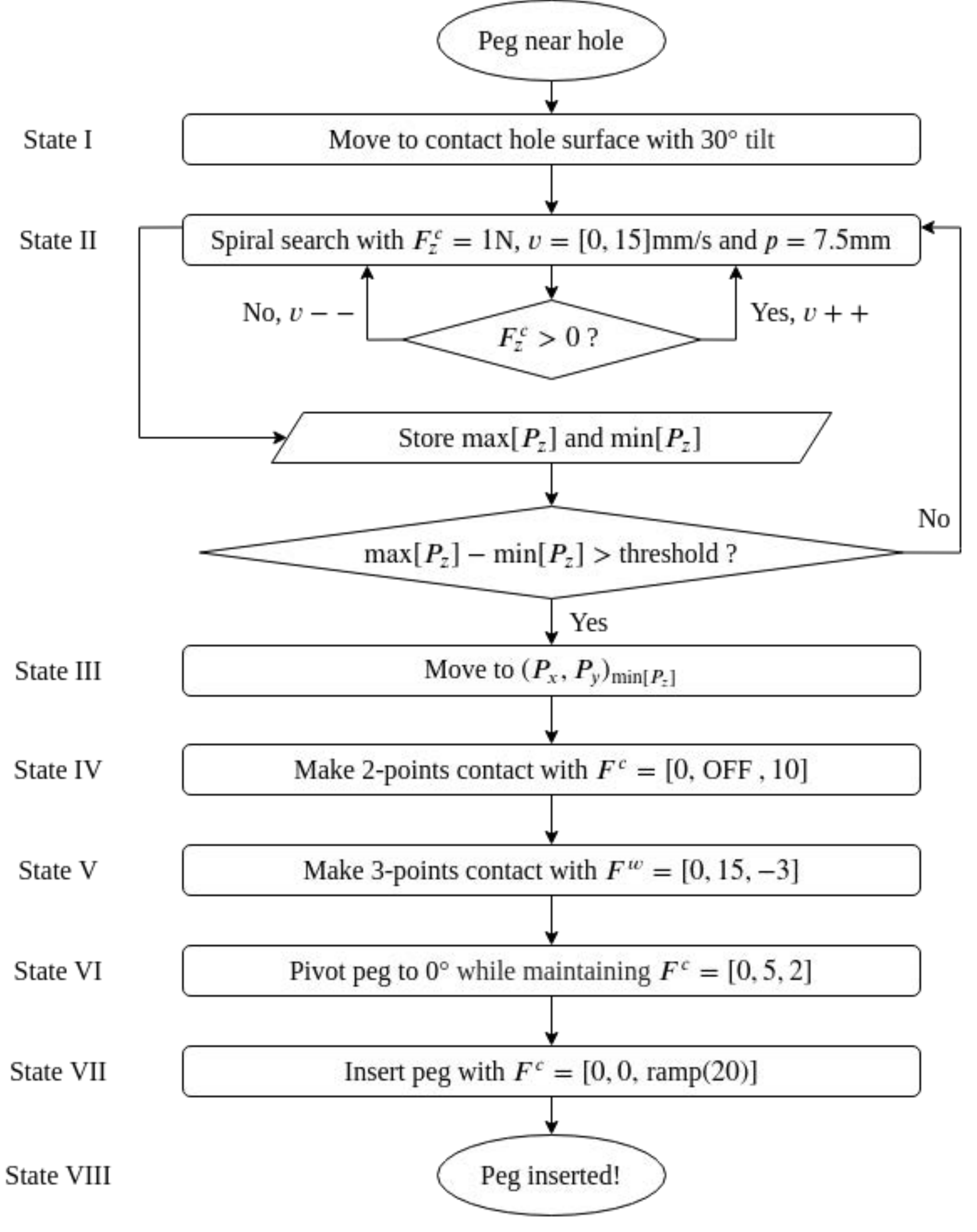}
\caption{Flowchart of assembly strategy}
\label{fig: flowchart assembly}
\end{center}
\end{figure}

Here we present our high-level assembly strategy, which can be broken
down into two phases of search and insertion as first proposed in
\cite{inoue1974force}, see illustration in Fig.~\ref{fig: search and
  insertion} and flowchart in Fig.~\ref{fig: flowchart assembly}.

\begin{figure*}
\begin{center}
\includegraphics[width = 6.8in]{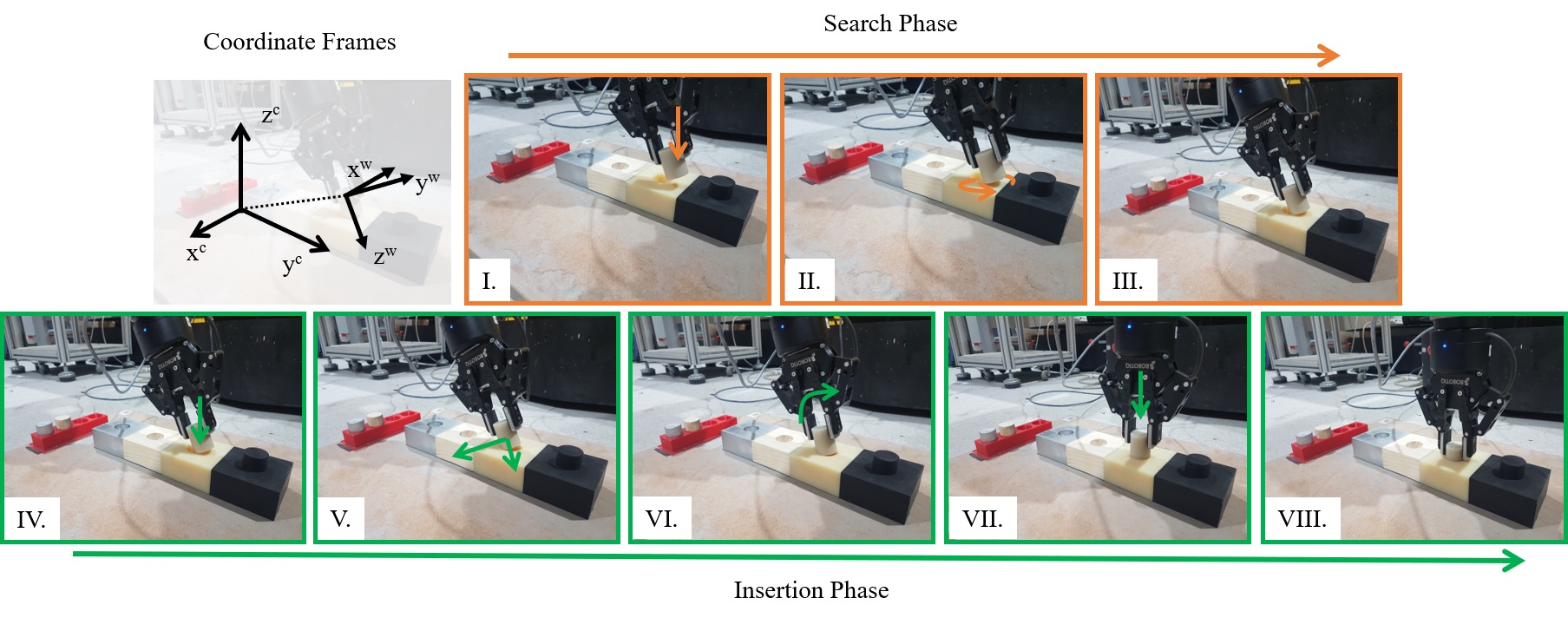}
\caption{Search and insertion phases and individual states}
\label{fig: search and insertion}
\end{center}
\end{figure*}

While the core of this high-level strategy is adapted from the
literature, we propose two main innovations: (i) in the search phase,
we propose a modulation of the sliding speed to mitigate the
slip-stick friction on soft materials; (ii) in the insertion phase, we
propose an explicit decomposition according to the number of point
contacts, and use a different force set-point for each sub-phase.

In the sequel, we define $\mathbf{P} = [P_x, P_y, P_z]$,
$\mathbf{F^c} = [F^c_x, F^c_y, F^c_z]$, and
$\mathbf{F^w} = [F^w_x, F^w_y, F^w_z]$ as the position vector,
Cartesian force vector, and wrench force vector respectively.

\subsection{Search phase}
The search phase encompasses states I-III. In state I, the peg is
lowered to the hole surface with a $\ang{30}$ tilt. This tilt enables
further interpretation of force measurements and allow for greater
changes in height along the search surface
\cite{inoue1974force,lozano1984automatic}. The point contact also
increases the speed of hole localization, especially for chamferless
peg-in-hole assemblies \cite{inoue1974force}. Additionally, the tilt
increases the cross-sectional area along the bending axis of the peg,
reducing the severity of deformation and slip-stick in state II.

In state II, a hybrid position-force controller is utilized to perform
an extensive spiral search, where the z-axis is controlled by a force
controller while the planar x-y axes are position-controlled to trace
an Archimedean spiral. The continuous spiral search path promises
efficiency and robust coverage of the search space
\cite{chhatpar2001search}. We first define $(P_{x_0}, P_{y_0})$ as the
initial position coordinates and let $v$ and $p = 7.5 \si{mm}$ be the
velocity and pitch of search path respectively. Then, with a small
$\Delta t = 1/125$, we can use Taylor's first order approximation to
obtain the estimated polar coordinates $\theta_{\mathrm{est}}$ and $r$. With
$r_0 = 0$ and $\theta_0 = 0$, the subsequent $P_x$ and $P_y$ position
commands for spiral search are defined as:

$$ \theta_{\mathrm{est}} = \sum \Delta \theta \hspace{0.5mm}, \hspace{2mm} \Delta\theta = \frac{v}{\sqrt{r^2 + \frac{p^2}{4 \pi^2}}}  \Delta t;$$
$$ r = \frac{p}{2 \pi} \theta_{\mathrm{est}}, $$
$$ P_x = P_{x_0} + r\cos(\theta_{\mathrm{est}}), $$
$$ P_y = P_{y_0} + r\cos(\theta_{\mathrm{est}}). $$

We can determine the position of the hole when the peg dips
significantly into the search plane. It is thus ideal for the force
controller to maintain $F_z^c > 0$ during the search. Also, to reduce
the effects of slip-stick friction along the search plane, it is
necessary for $F_z^c$ reference to kept small. Therefore, we command
the normal reference force as $F^c_z = 1.0 \si{N}$.  Yet, a lower
$F_z^c$ reference induces a slower change along the z-axis when the
peg is above the hole. In order to maintain $F_z^c > 0$ even when
above the hole, the search velocity should be constrained. Without
compromising on the search efficiency outside the hole, we can
decelerate $v$ to $0 \si{mm/s}$ when $F_z^c = 0$, and accelerate it to
a maximum of $15 \si{mm/s}$ when $F_z^c > 0$.

We continue to store the $\max[P_z]$ and $\min[P_z]$ detected during
the spiral search. Upon detection of a pre-defined difference in
$\max[P_z]$ and $\min[P_z]$, the hole is located and the peg is
moved linearly towards $(P_x, P_y)_{\min[P_z]}$ in state III. Such an
algorithm ensures robustness in hole detection for all possible
initial contact points. Note that the threshold should be sufficiently
large to account for any possible noise in sensor measurements or
slight displacements in the setup. In theory, the threshold and pitch
of the spiral search is related by the area covered within the
isolines of the tilted peg depth profile. An oversized threshold will
require a reduced pitch during the spiral search for successful hole
localization.

\subsection{Insertion phase}
For the insertion phase entailing states IV-VIII, all three axes
switch to active force control. To ensure robustness in the insertion,
each subsequent state is only fired after a steady state is attained
in the measurements. Due to 2-finger clamping of the peg, the peg
occasionally pivots about the contact point along the x-axis, thereby
reducing the reliability of force measurements and control along that
axis. Thus, the constraint for steady state along the x-axis is
relaxed and the control along the x-axis is switched off once steady
state is attained in subsequent states.

We first attempt to bring the peg to establish a stable and consistent
contact with the hole. For that, we command the tilted peg to make
2-points contact at $P_y$ by utilizing compliant force control with
$\mathbf{F^c} = [0, \mathrm{OFF}, 10]$ in state IV. The peg will trace
the edge of the hole to slide to its lowermost point at a fixed $P_y$ to
establish 2-points contact. In our setup, a 2-points contact implies
that the peg is centralized along the x-axis, and the hole center
subsequently lies along the y-axis\cite{sharma2013intelligent}. Next
in state V, we aim to establish a stable 3-points contact to increase
the robustness of our insertion strategy. By tracking forces and
moving in the wrench coordinate frame, $\mathbf{F^w} = [0, 15, -3]$,
we increase the likelihood of the peg remaining within the hole during
motion.

After attaining 3-points contact, we proceed to align and insert the
peg. In state VI, we conduct a position-controlled tilting of the peg
by pivoting and maintaining contact of $\mathbf{F^c} = [0, 5, 2]$ with
the edge of the hole. Here, the fast performance of the force
controller is especially important to prevent excessive force
overshoot. Then, we center and insert the peg with $\mathbf{F^c} = [0,
  0, \mathrm{ramp}(20)]$ in state VII. The ramp input mimics the
careful insertion by humans and allows time to correct minor
misalignments. Finally, the assembly is completed at state VIII.

\section{Experiments}
\label{sec:experiments}

We evaluate the performance of the low-level force controllers
expressed in Section \ref{sec:controller} for peg-in-hole assembly
using the high-level assembly strategy described in Section
\ref{sec:strategy}. For that, we conduct two different sets of
experiments with the CCS and PID force controllers.

\subsection{Experimental setup}
The assembly experiments are performed on a 6-axis position-controlled
Denso VS-087 robotic arm. Its 2-fingered end-effector gripper is
equipped with an ATI Force/Torque sensor, communicating at $125
\si{Hz}$ with an Ubuntu 16.04 computer. Chamferless pegs and insertion
blocks are manufactured to an ISO 30 H9d9 fit, with a clearance of
$0.117 \pm 0.052 \si{mm}$ using four different materials. The height
of the peg and hole is $40 \si{mm}$ and $20 \si{mm}$ respectively. The
materials and corresponding overall contact stiffness are shown in
Table \ref{table: material stiffnesses}.

\begin{table}[htp]
\caption{Materials and corresponding stiffness values}
\label{table: material stiffnesses}
\begin{center}
\begin{tabular}{|l|c|}
\hline
Material & Stiffness (\si{N/mm}) \\ \hline
Shore 30A silicone rubber & 10 \\ \hline
ABS plastic & 50 \\ \hline
Pine wood & 65 \\ \hline
Aluminum & 100 \\ \hline
\end{tabular}
\end{center}
\end{table}

We subsequently design three force controllers: CCS, INT$_\mathrm{s}$
(INTegral controller optimized for \underline{s}oft contacts), and
INT$_\mathrm{h}$ (INTegral controller optimized for \underline{h}ard
contacts).
\begin{itemize}
\item CCS force controller is designed using the CCS framework for a
  first order time response of $0.17 \si{s}$ with a nominal stiffness
  of $10 \si{N/mm}$ and robust stability against contact stiffnesses
  up to $100 \si{N/mm}$;
\item INT$_\mathrm{s}$ is an integral force controller tuned with the
  same time response as the CCS controller at the nominal
  $10 \si{N/mm}$ stiffness;
\item INT$_\mathrm{h}$ is an integral force controller tuned to be
  stable against contact stiffnesses up to $100 \si{N/mm}$.
\end{itemize}

For the two experiments, we define the nominal contact point as where
the lowest point of the peg coincides with the center of the hole at
the end of state I. We evaluate the performance of the different force
controllers with reference to this nominal contact point.

\subsection{Experiment 1: CCS force controller}
\label{subsection: experiment 1}

In our first experiment, we utilize the CCS force controller as the
low-level force controller for assembly. We aim to show that our
controller achieves robust insertion for random initial contact points
in all four materials.

\begin{figure}[htp]
  \begin{center}
    \subfloat[Silicone rubber\label{fig: ccs rubber}]{%
      \includegraphics[width=3.4in]{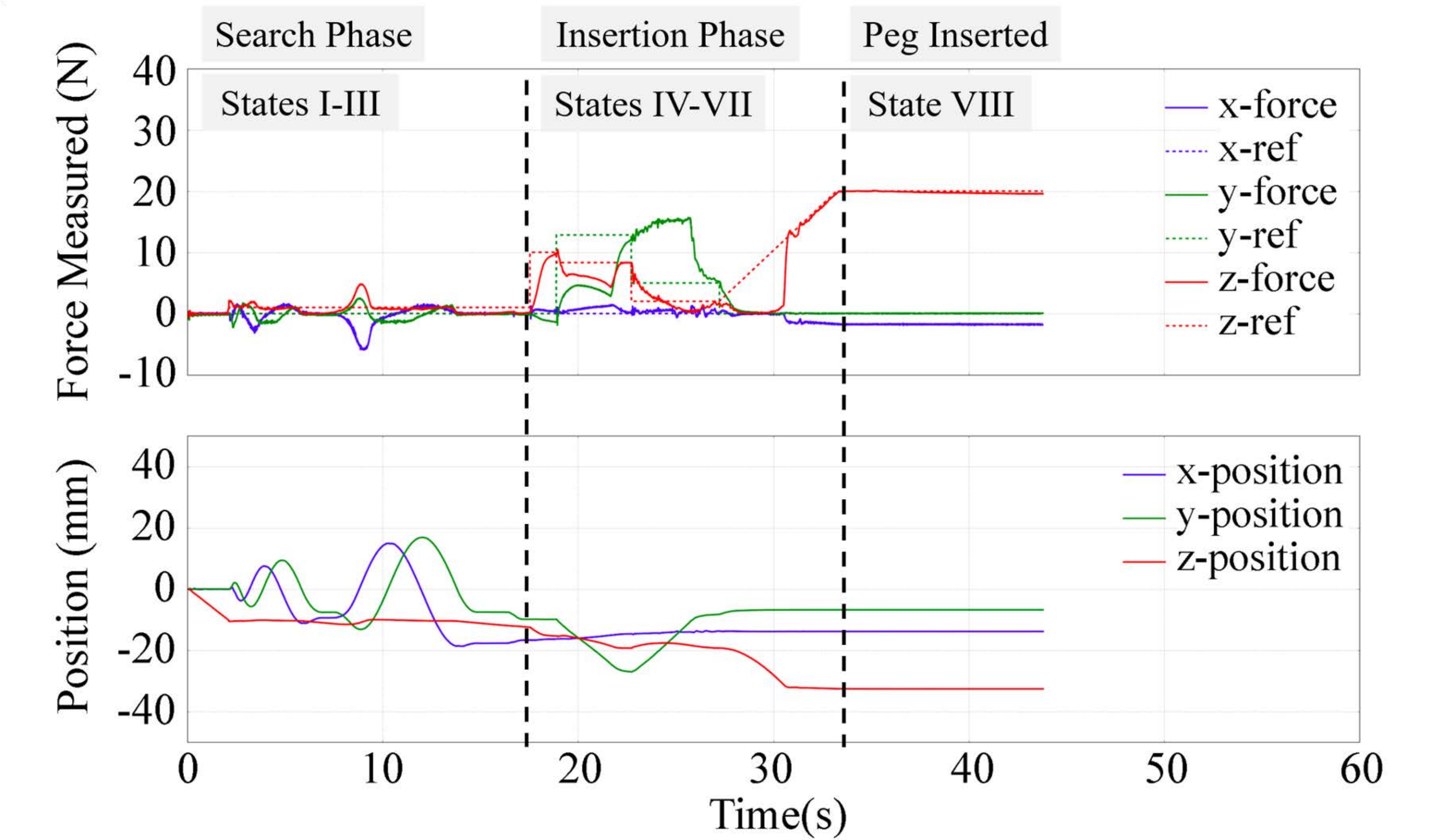}}
    \\
    \subfloat[ABS plastic\label{fig: ccs plastic}]{%
      \includegraphics[width=3.4in]{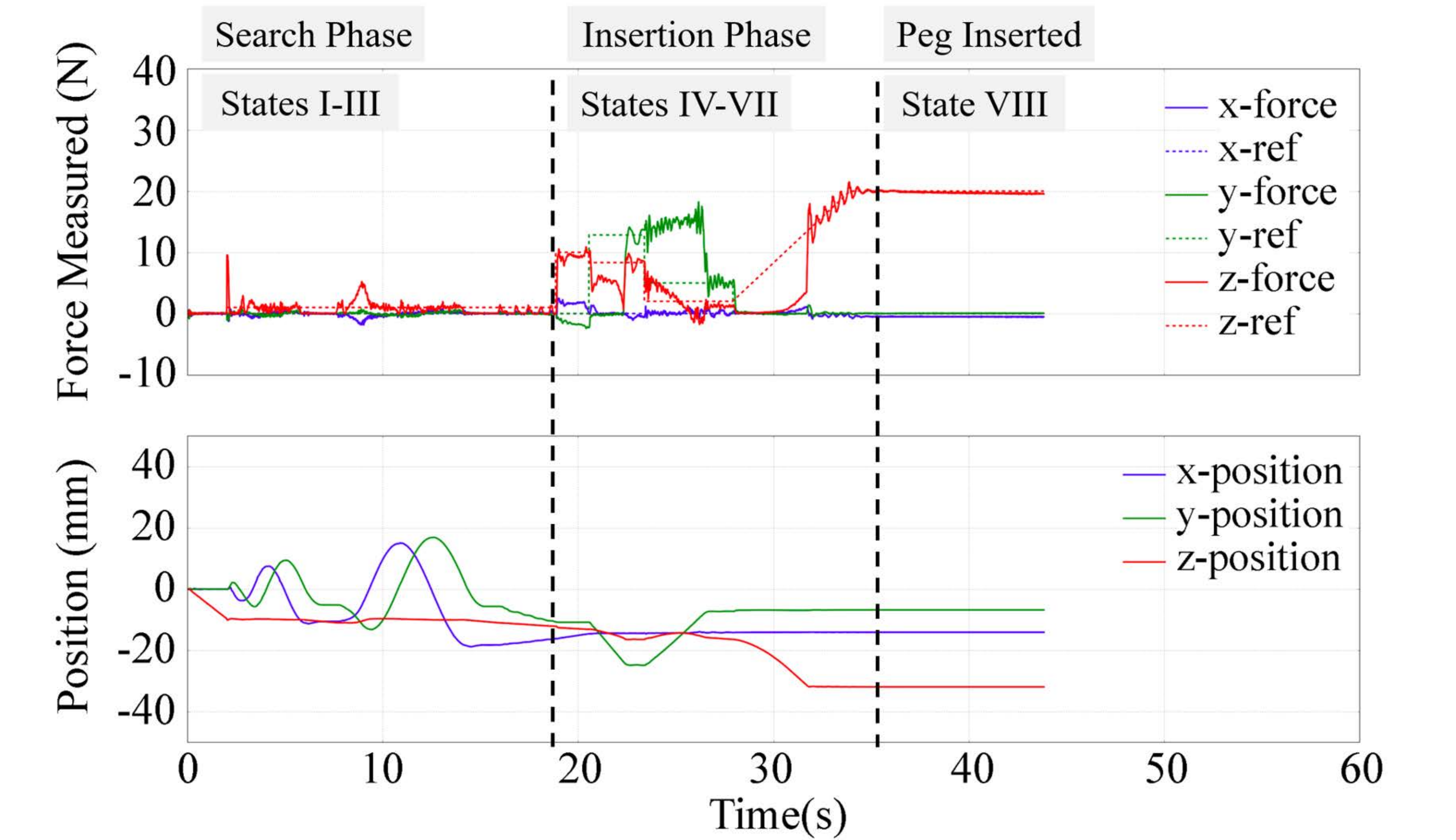}}
    \\
    \subfloat[Pine wood\label{fig: ccs wood}]{%
      \includegraphics[width=3.4in]{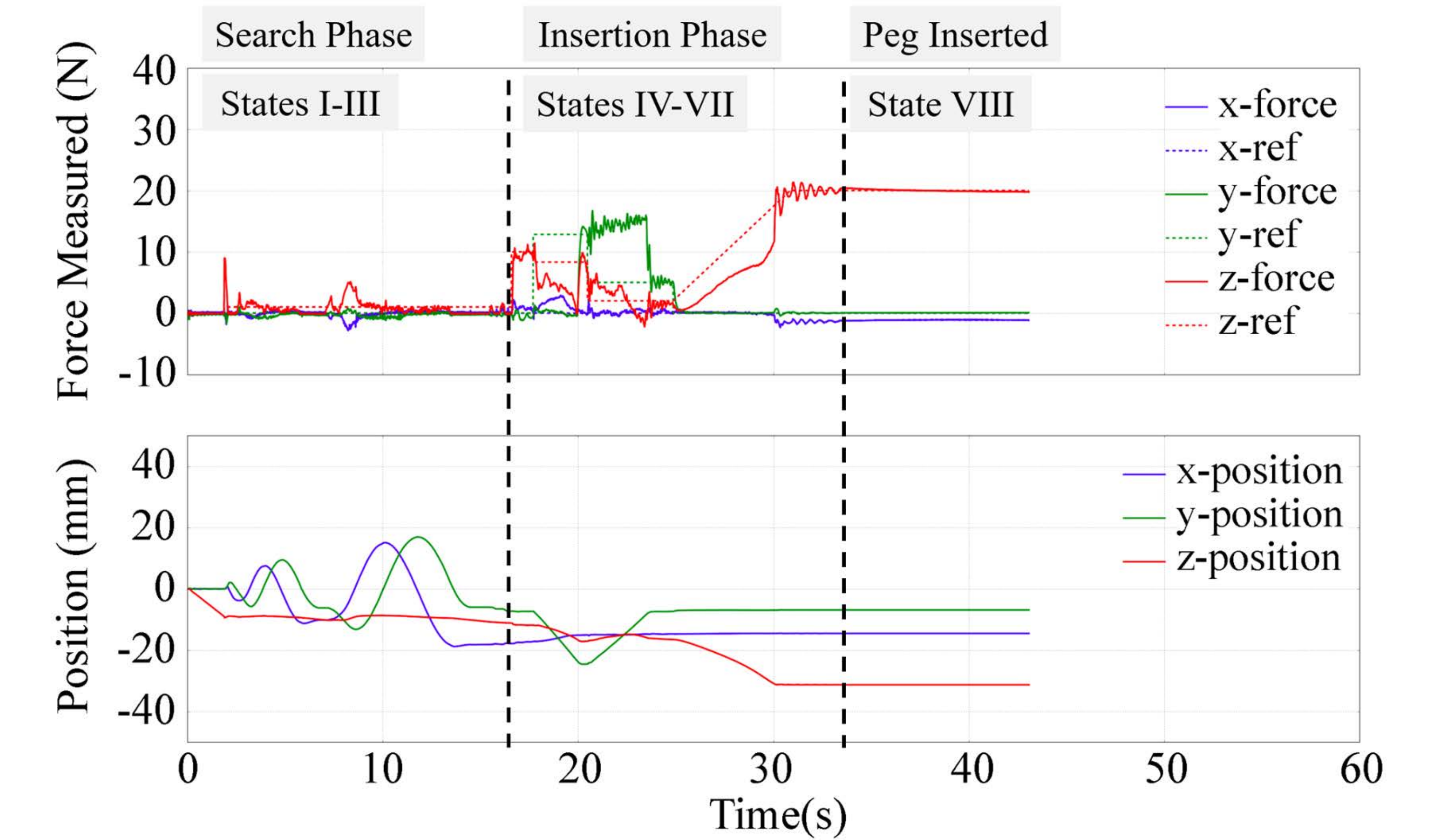}}
    \\
    \subfloat[Aluminum\label{fig: ccs aluminum}]{%
      \includegraphics[width=3.4in]{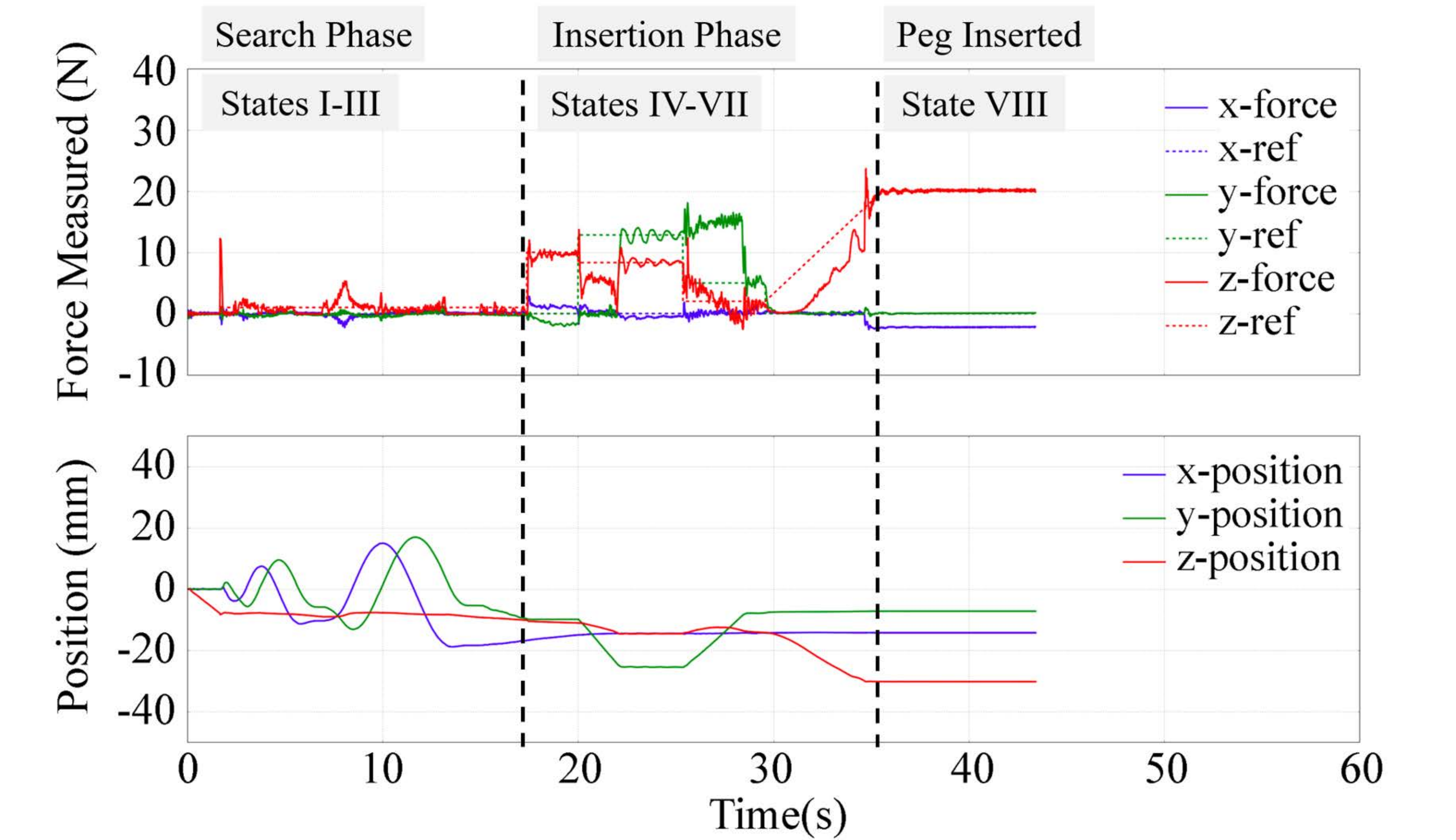}}
    \caption{Force and position measurements in the peg-in-hole
      experiments performed using CCS. Four materials were tested,
      with increasing stiffnesses: (a) silicone rubber, (b) ABS
      plastic, (c) pine wood, and (d) aluminum. See video of the
      experiment at
      \url{https://youtu.be/dgmsPGvF3d0}}
    \label{fig:ccs} 
  \end{center}
\end{figure}

We perform a total of 48 trials -- 12 on each material -- with
starting positions randomly sampled within a $2 \si{cm}$ radius from
the nominal contact point. Using the CCS force controller, we achieve
robust assembly performance with a $100\%$ success rate, as shown in
Table \ref{tab:results}. The average and maximum assembly durations
after initial contact are $26.9 \si{s}$ and $38 \si{s}$ respectively.
See also the video of the experiment at
\url{https://youtu.be/dgmsPGvF3d0}, as well as the
force and position measurements in Fig.~\ref{fig:ccs}.

\begin{table}[htp]
  \caption{Experiment results for CCS controller}
  \label{tab:results}
  \begin{center}
    \begin{tabular}{|m{2.5cm}|C{1.1cm}|C{1.1cm}|C{1.1cm}|}
      \hline
      Material & Total no. of tests & Success rate (\%) & Average time (\si{s}) \\\hline
      30A silicone rubber & 12 & 100 & 26.2\\ \hline
      ABS plastic & 12 & 100 & 28.3 \\ \hline
      Pine wood & 12 & 100 & 25.4 \\ \hline
      Aluminum & 12 & 100 & 27.8 \\ \hline
      Total & 48 & 100 & 26.9 \\ \hline
    \end{tabular}
  \end{center}
\end{table}

\subsection{Experiment 2: PID force controllers}

For our second experiment, we utilize the two integral force
controllers individually as the low-level force controller for
assembly.

\begin{figure}[htp]
  \begin{center}
    \includegraphics[width = 3.4in]{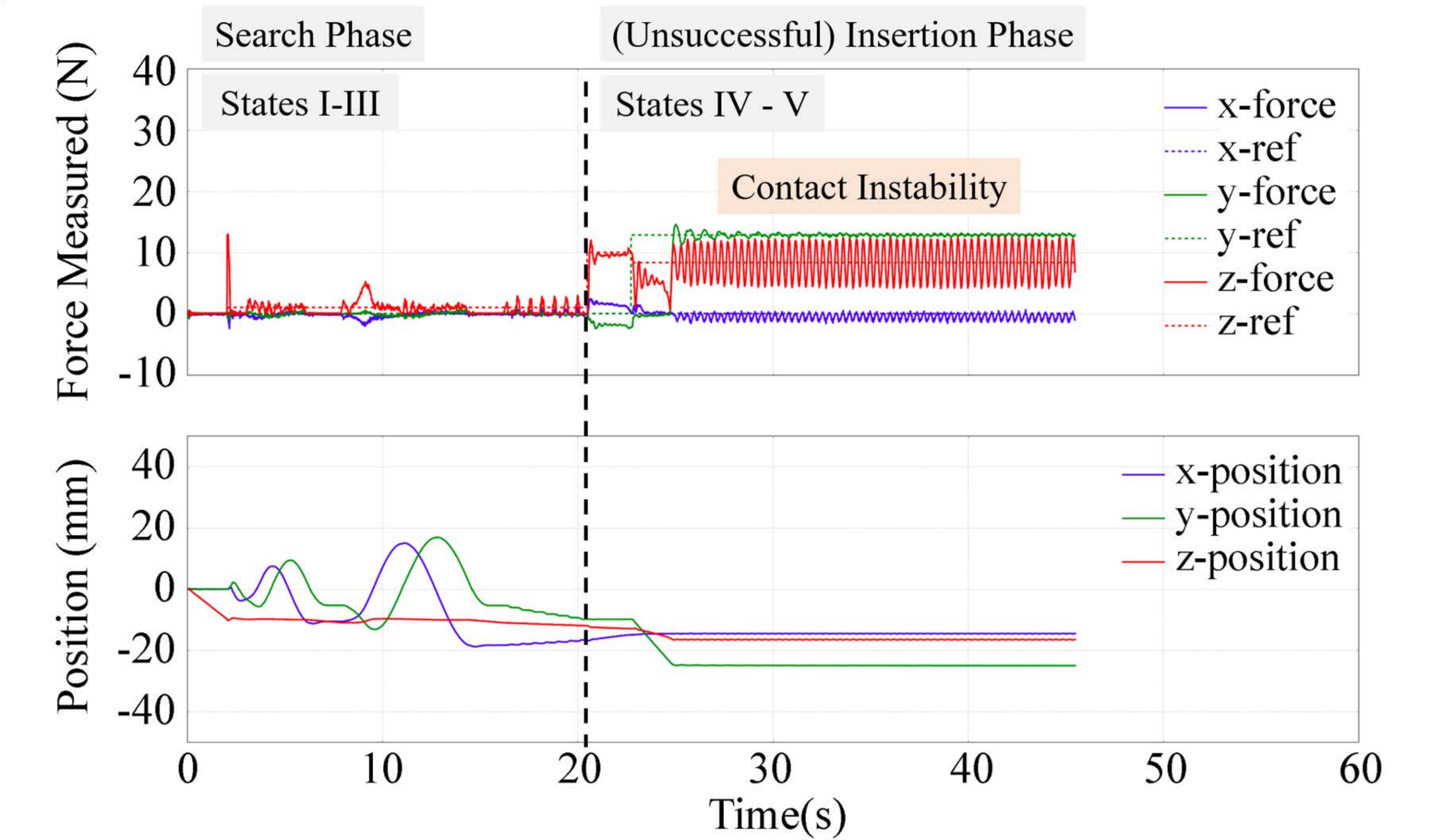}
    \caption{Force and position measurements in the experiment
      performed using INT$_\mathrm{s}$ (optimized for rubber) on
      plastic surface (which is harder than rubber). Note that the
      robot goes into instability at state V, around $t = 24 \si{s}$.}
    \label{fig: PIss plastic}
  \end{center}
\end{figure}

The experiments using the INT$_\mathrm{s}$ force controller show that
while the INT$_\mathrm{s}$ force controller is viable for the
$10 \si{N/mm}$ silicone rubber environment, the assembly task cannot
be completed with stiffer contacts. Fig. \ref{fig: PIss plastic}
demonstrates the INT$_\mathrm{s}$ force controller going into
instability at $t = 25 \si{s}$ during the insertion phase of the ABS
plastic assembly. Similar robustness issues are found during the pine
wood and aluminum assembly.

Using the INT$_\mathrm{h}$ force controller, the assembly also fails as the
force controller reacts slower to changes in forces, causing a longer
settling time and excessive force overshoot. Fig. \ref{fig: PIstiff
  wood} shows that the INT$_\mathrm{h}$ force controller has a slower response
time in inducing changes in $P_z$ when the peg is above the hole. As
such, using the INT$_\mathrm{h}$ controller takes approximately $50\%$ longer
than using the CCS controller for search after contact. It also takes
approximately $50\%$ longer to establish stable 3-points contact as
compared to the CCS controller during insertion. Finally, with the
position-commanded tilt, the assembly experiences excessive force
overshoot $ |\mathbf{F^c}| > 50 \si{N}$ and is halted to avoid
damaging parts.

\begin{figure}[htp]
  \begin{center}
    \includegraphics[width = 3.4in]{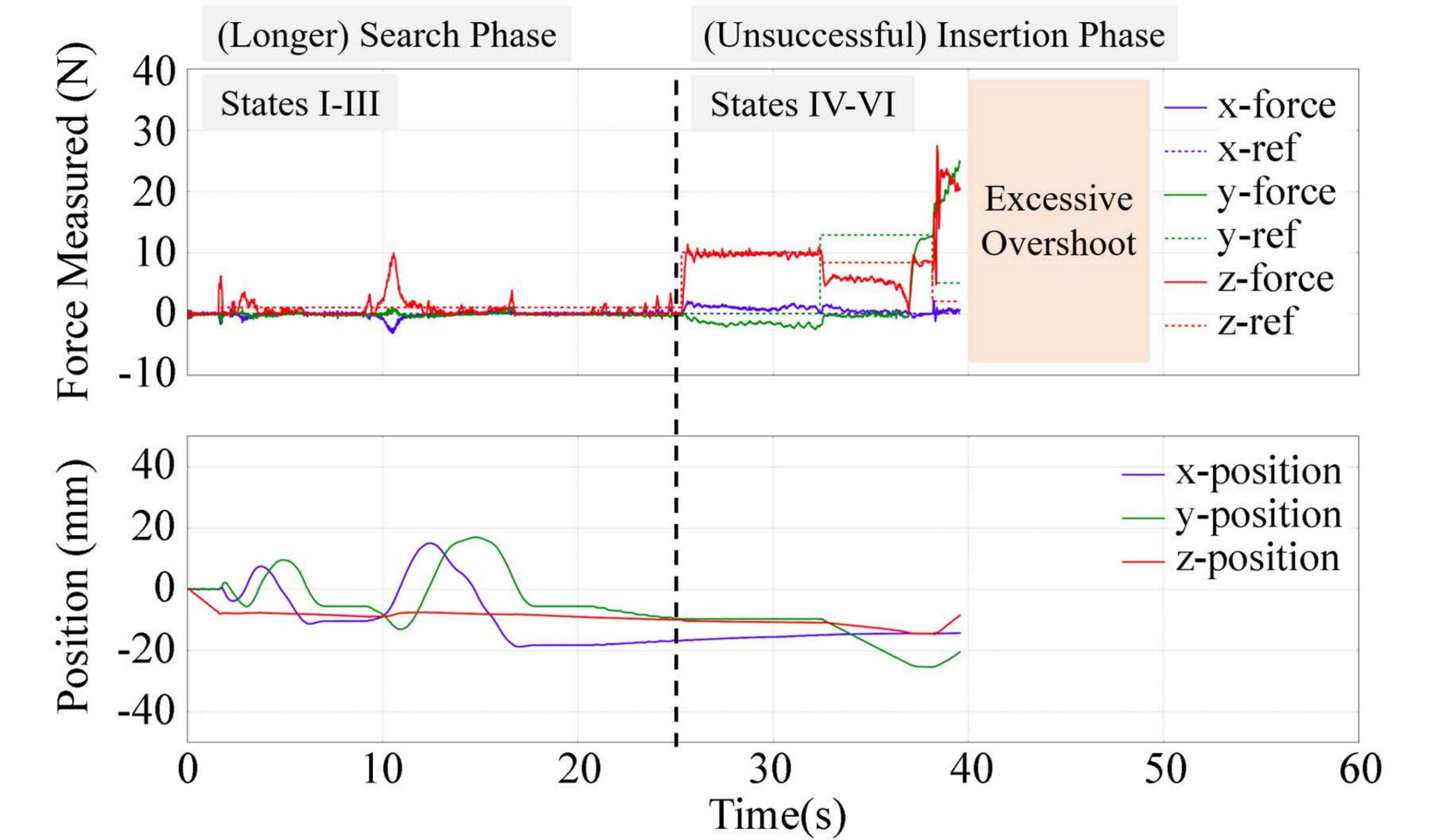}
    \caption{Force and position measurements in the experiment
      performed using INT$_\mathrm{h}$ (optimized for aluminum) on
      aluminum surface. Note the large force overshoot ($>50\si{N}$)
      caused by the sluggishness of the controller at State VI, around
      $t = 40 \si{s}$.}
    \label{fig: PIstiff wood}
  \end{center}
\end{figure}

\subsection{Discussions}

Our results highlight the limitations of conventional PID controllers
in performing peg-in-hole assembly for environments of different
stiffnesses. An integral force controller tuned for good nominal
performance with silicone rubber goes into instability during
peg-in-hole assembly in environments with higher contact stiffnesses
(plastic, wood, aliminum). At the other end of the spectrum, an
integral force controller tuned for stable contact with aluminum
experiences a long settling time and large force overshoots during the
assembly.

Meanwhile, the proposed CCS force controller tuned for good nominal
performance in silicone rubber is able to perform peg-in-hole assembly
similarly well in all four materials of different contact stiffnesses,
without any gain tuning. It also performs better with shorter settling
time and smaller force overshoot than a robustly-tuned integral force
controller. Furthermore, the assembly strategy utilizing the CCS force
controller also demonstrates high robustness to different initial
contact points.

\section{Conclusions}
\label{sec:conclusions}

In this paper, we build upon the recently-developed Convex Controller
Synthesis (CCS)~\cite{pham2019convex} to enable high-precision
assembly across a wide range of surface stiffnesses \emph{without any
  surface-dependent tuning}. Specifically, we have demonstrated
peg-in-hole assembly with \SI{100}{\micro\metre} clearance, initial
position uncertainties up to \SI{2}{cm}, and for four types of peg and
hole materials -- rubber, plastic, wood, aluminum -- whose stiffnesses
range from 10 to \SI{100}{N/mm}, using a single controller.

Our results highlight the performance and robustness of Convex
Controller Synthesis over classical PID schemes. In future work, we
shall explore the combination of CCS with Reinforcement Learning to
generate versatile and robust assembly strategies.



\section*{ACKNOWLEDGMENT}

Lee Y. J. W. thanks Hung Pham and Zhang Xu for their guidance.


\bibliography{pihreferences}

\end{document}